\documentclass[conference]{IEEEtran}
\IEEEoverridecommandlockouts

\usepackage{cite}
\usepackage{amsmath,amssymb,amsfonts}
\usepackage{algorithmic}
\usepackage{graphicx}
\usepackage{textcomp}
\usepackage{xcolor}
\usepackage{multirow}
\usepackage{multirow}
\usepackage{makecell}
\usepackage{booktabs} 
\usepackage{multirow} 

\usepackage{bbding}   
\usepackage{array}

\def\BibTeX{{\rm B\kern-.05em{\sc i\kern-.025em b}\kern-.08em
    T\kern-.1667em\lower.7ex\hbox{E}\kern-.125emX}}
\begin{document}





\renewcommand{\thefootnote}{\arabic{footnote}}

\makeatletter
\def\IEEEauthorrefmark#1{\textsuperscript{\footnotesize #1}}
\makeatother

\newcommand{\corrauthor}{\textsuperscript{\dag}}

\title{EndoWave: 4D Gaussian Splatting with Rational Wavelet for Endoscopic Reconstruction
\thanks{\noindent\parbox{\linewidth}{%
\corrauthor~Corresponding authors.}}%
}

\author{%
  \IEEEauthorblockN{%
    Taoyu Wu\IEEEauthorrefmark{1,3},
    Yiyi Miao\IEEEauthorrefmark{2,4},
    Jiaxin Guo\IEEEauthorrefmark{5},
    Ziyan Chen\IEEEauthorrefmark{1},
    Sihang Zhao\IEEEauthorrefmark{1}, \\
    Zhuoxiao Li\IEEEauthorrefmark{6}, 
    Zhe Tang\IEEEauthorrefmark{7},
    Baoru Huang\IEEEauthorrefmark{4}\corrauthor,
    Limin Yu\IEEEauthorrefmark{1}\corrauthor}
  \IEEEauthorblockA{%
    \IEEEauthorrefmark{1}School of Advanced Technology, Xi'an Jiaotong-Liverpool University, China\\
    \IEEEauthorrefmark{2}School of AI and Advanced Computing, Xi'an Jiaotong-Liverpool University, China\\
    \IEEEauthorrefmark{3}School of Physical Sciences, University of Liverpool, Liverpool, United Kingdom\\
    \IEEEauthorrefmark{4}School of Computer Science and Informatics, University of Liverpool, United Kingdom\\
    \IEEEauthorrefmark{5} Department of Mechanical and Automation Engineering, The Chinese University of Hong Kong, Hong Kong, China\\
    \IEEEauthorrefmark{6}Urban Governance and Design Thrust, The Hong Kong University of Science and Technology (Guangzhou), China\\
    \IEEEauthorrefmark{7}Institute of Artificial Intelligence Innovation, Zhejiang University of Technology, China}
}

\maketitle

\begin{abstract}
In robot-assisted minimally invasive surgery, accurate 3D reconstruction from endoscopic video is vital for downstream tasks and improved outcomes. However, endoscopic scenarios present unique challenges, including photometric inconsistencies, non-rigid tissue motion, and view-dependent highlights. Most 3DGS-based methods that rely solely on appearance constraints for optimizing 3DGS are often insufficient in this context, as these dynamic visual artifacts can mislead the optimization process and lead to inaccurate reconstructions. To address these limitations, we present \textbf{EndoWave}, a unified spatio-temporal Gaussian Splatting framework by incorporating an optical flow-based geometric constraint and a multi-resolution rational wavelet supervision. First, we adopt a unified spatio-temporal Gaussian representation that directly optimizes primitives in a 4D domain. Second, we propose a geometric constraint derived from optical flow to enhance temporal coherence and effectively constrain the 3D structure of the scene. Third, we propose a multi-resolution rational orthogonal wavelet as a constraint, which can effectively separate the details of the endoscope and enhance the rendering performance. Extensive evaluations on two real surgical datasets, EndoNeRF~\cite{endonerf} and StereoMIS~\cite{stereomis}, demonstrate that our method \textbf{EndoWave} achieves state-of-the-art reconstruction quality and visual accuracy compared to the baseline method.
\end{abstract}


\section{Introduction}

Three-dimensional reconstruction in endoscopy is a crucial technology for enhancing surgeon perception and guidance during minimally invasive surgery. After collecting multi-frame endoscopic video, reconstructing the operative scene improves spatial understanding and supports downstream tasks~\cite{xu2024review}. In robotic-assisted procedures, accurate modelling of dynamic anatomy benefits surgical planning and intraoperative navigation. However, the endoscopic environment introduces unique challenges for vision algorithms. The confined field of view, frequent occlusions such as instruments or tissue folds, specular highlights, and continuous non-rigid deformations undermine the static-scene assumptions and dense feature correspondences required by conventional stereo, Simultaneous Localization and Mapping (SLAM), and Structure-from-Motion pipelines~\cite{wang2024endogslam}. These factors directly violate the core assumptions of static scenery and abundant feature correspondences upon which conventional 3D mapping approaches are built, often leading to incomplete or inaccurate reconstructions. To overcome these fundamental limitations, the field has increasingly turned towards learning-based methods capable of modeling complex, dynamic scenes.

\begin{figure}[!t]
    \centering
    \includegraphics[width=\linewidth]{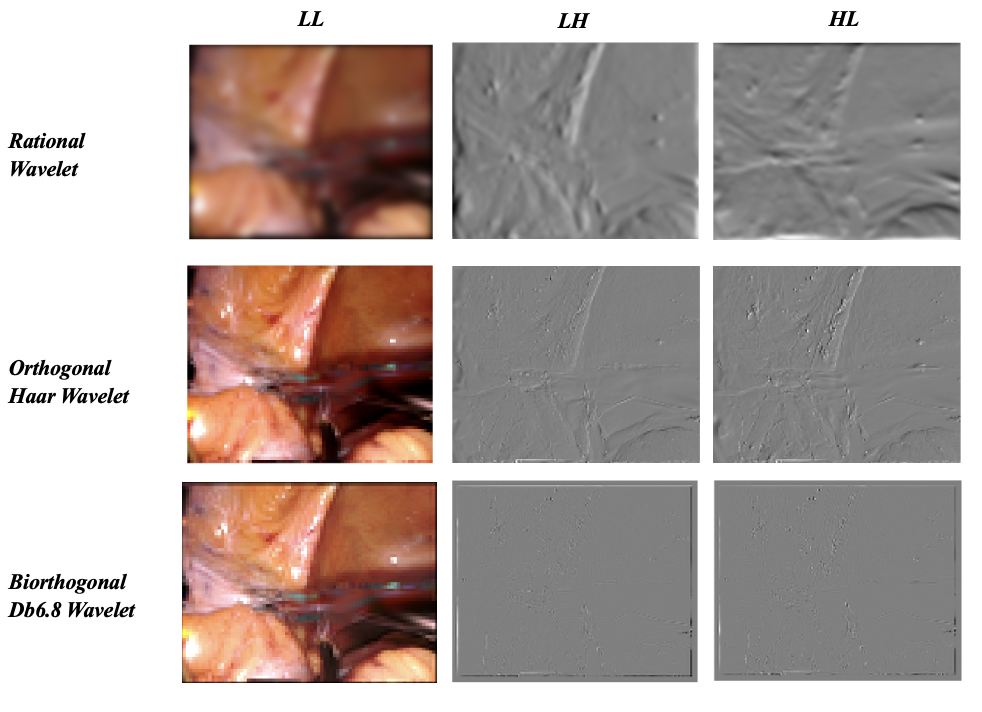}
    \vspace{-2em}
    \caption{\textbf{Wavelet decomposition visualization on EndoNeRF dataset.} We propose a rational wavelet decomposition that is effective for endoscopic scenarios. 
    Our proposed rational wavelet is compared against the standard Orthogonal Haar and Biorthogonal bior6.8, visualizing their respective LL, LH, and HL components.}
    \label{fig:wavelet}
\end{figure}

Among these learning-based solutions, Neural Radiance Fields (NeRF)~\cite{mildenhall2021nerf,guo2024uc} have emerged as a paradigm, demonstrating exceptional fidelity in novel view synthesis and 3D reconstruction by representing scenes implicitly with neural networks. Early adaptations of this method to endoscopy, such as EndoNeRF~\cite{endonerf}, introduced dual-field representations to model deformable tissue through a canonical radiance field paired with a time-varying deformation field. Subsequent methods sought to refine this approach; for instance, EndoSurf~\cite{zha2023endosurf} employed a neural signed distance function to explicitly enforce surface consistency, thereby improving geometric accuracy for dynamic tissues. Despite their rendering quality, NeRF-based methods are still hindered by slow inference and extensive training requirements, which make them impractical for real-time surgical use~\cite{freesurgs, wu2025endoflow}. Moreover, purely photometric NeRF optimizations can struggle with localization accuracy and temporal consistency in monocular endoscopy scenarios.

Recently, 3D Gaussian Splatting (3DGS) has emerged as a highly efficient alternative for real-time scene representation and rendering~\cite{kerbl20233dgs,ulsr}. In contrast to the implicit neural fields of NeRF, 3DGS models a scene as an explicit collection of anisotropic 3D Gaussians, each defined by properties like position, orientation, color, and opacity. This explicit structure permits rasterization at extremely high frame rates, making it a compelling candidate for surgical applications. Initial efforts to adapt 3DGS to the surgical domain, EndoGaussian~\cite{liu2024endogaussian} integrated depth priors to achieve near real-time reconstruction of deformable anatomy. Building on this, subsequent research has focused on extending 3DGS to model dynamic, 4D scenes. A foundational 4D Gaussian Splatting framework~\cite{wu4dgs} proposed a holistic spatio-temporal representation, where a canonical set of Gaussians is deformed over time using a lightweight neural network. 
Deform3DGS~\cite{yang2024deform3dgs} further refines the concept with a flexible deformation model that tracks tissue motion. Endo-4DGS~\cite{huang2024endo} couples 3DGS with a learned deformation field to achieve monocular reconstruction. SurgicalGS~\cite{chen2025surgicalgs} refines the reconstruction by fusing multi-frame depth cues and imposing geometric constraints to better capture fine anatomical details.

Despite this progress, significant gaps remain for 4D reconstruction in challenging endoscopic scenes. First, many dynamic 3DGS methods rely on a two-stage paradigm that first learns a static canonical representation and then trains a separate network to model its deformation over time. This separation can be suboptimal for complex non-rigid motion and complicates the optimization process. Second, existing approaches are predominantly guided by photometric consistency, lacking explicit constraints to ensure that the reconstructed motion is geometrically consistent with the observed pixel-level dynamics. This can lead to temporal artifacts and inaccurate deformation tracking. Third, the unique visual characteristics of endoscopic scenes, which contain smooth, low-frequency surfaces and sharp, high-frequency specular highlights, are difficult for standard loss functions to model accurately, often resulting in blurred details or noisy artifacts~\cite{guo2022review}.

To address these limitations, we present \textbf{EndoWave}, a 4D Gaussian Splatting (4DGS) framework guided by optical flow and multi-resolution rational wavelets for high-fidelity reconstruction of dynamic endoscopic scenes. First, we adopt a unified temporal representation that directly optimizes Gaussians in a 4D spatio-temporal domain, avoiding the need for a canonical model and a separate deformation network. This approach inherently captures complex tissue motion and simplifies the training pipeline. Second, we introduce a geometric constraint by leveraging optical flow. We enforce consistency between the projected 2D motion of our 4D Gaussians and the optical flow estimated by off-the-shelf methods, such as GMFlow~\cite{xu2022gmflow}, ensuring that the reconstructed scene flow accurately reflects the observed pixel dynamics. Third, we propose a multi-resolution constraint using rational wavelets, which are better suited for the unique frequency characteristics of endoscopic imagery than traditional wavelets. By decomposing the rendered and ground-truth images into distinct frequency bands, we can preserve the global tissue structure while reconstructing high-frequency details, such as vessel boundaries and specular reflections. Extensive experiments on the EndoNeRF and StereoMIS datasets demonstrate that our method achieves state-of-the-art reconstruction quality, surpassing NeRF-based and 3DGS-based baselines in both geometric accuracy and visual fidelity, while maintaining interactive rendering rates.

Our contributions can be summarized in threefold:

\begin{itemize}
    \item Unified Spatio-Temporal Gaussian Representation. We represent the scenes as spatio-temporal Gaussians optimized directly over time, avoiding the conventional two-stage canonical and deformation pipeline.
    
    \item Flow-induced Geometric Constraint. We introduce an optical flow-derived loss to serve as a geometric constraint. 
    
    \item Multi-Resolution Wavelet Supervision. We propose a plug-in rational wavelet component that supervises the reconstruction across multiple frequency bands.

\end{itemize}

\begin{figure*}[!t]
    \centering
    \includegraphics[width=0.89\textwidth]{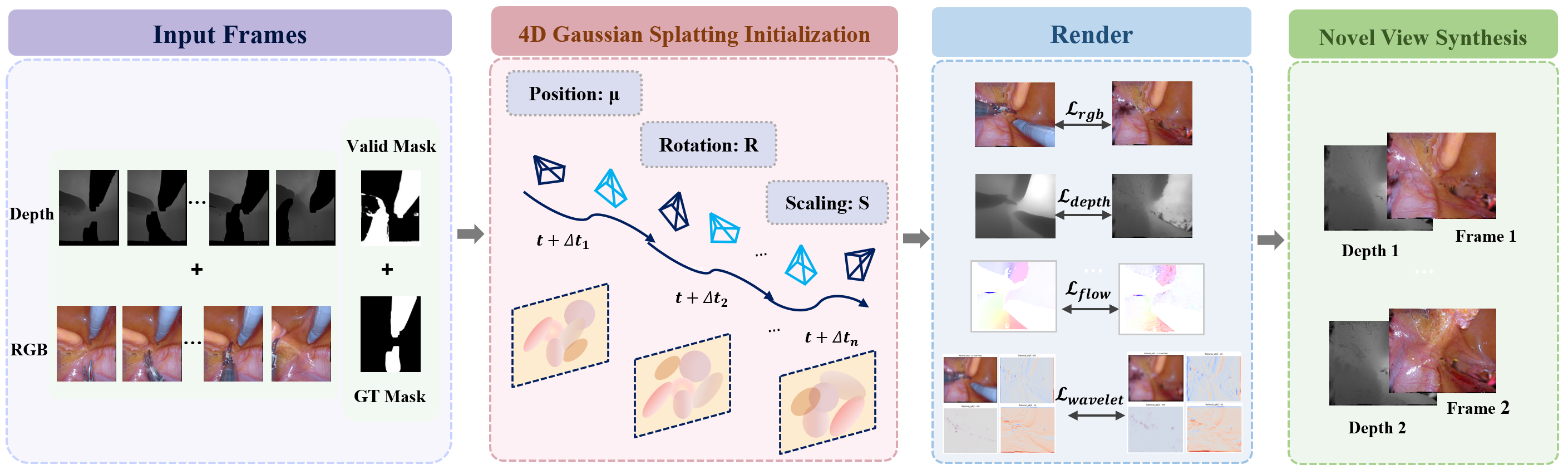}
    \caption{\textbf{Overall of the proposed framework.} We take RGB-D frames as input to initialize a set of 4D Gaussian Splatting primitives $\mathcal{G}$. Each primitive is decomposed into a conditional 3D Gaussian for its spatial representation and a marginal 1D Gaussian to model its temporal dynamics. The primitives are then jointly optimized using a composite loss function with RGB, depth, optical flow, and multi-scale wavelet supervision. After training, the model can render high-fidelity, time-evolving color and depth maps from novel viewpoints at any given time $t$.}
    \label{fig:pipeline}
\end{figure*}

\section{Related Work}

\subsection{Neural Radiance Fields in Surgical Reconstruction}
NeRF achieves impressive fidelity in novel view synthesis by learning a continuous function that maps 3D coordinates and view directions to color and density. Its adoption in medical imaging, while exploratory, has shown considerable promise~\cite{long2021dssr}. The pioneering work of EndoNeRF~\cite{endonerf} first adapted this paradigm to endoscopic environments, introducing a dual neural field approach to separately model tissue deformation and the canonical scene representation. Building upon this, EndoSurf~\cite{zha2023endosurf} enhanced geometric fidelity by incorporating signed distance functions and self-consistency constraints, which improved surface reconstruction accuracy at the cost of increased computational complexity. A primary challenge for these methods, however, is their computational intensity. To accelerate rendering for dynamic scenes, researchers have explored factorized or hybrid representations. Inspired by low-rank decomposition techniques, LerPlane~\cite{yang2023neural} accelerates the modeling of deformable tissue by introducing a 4D grid representation that factorizes the scene into compact spatial and temporal components. Despite these advances, NeRF-based methods still suffer from fundamental computational limitations. This performance gap has motivated the exploration of alternative, explicit representations that promise comparable quality with superior computational efficiency.

\vspace{-0.5em}

\subsection{3D Gaussian Splatting and Surgical Scene Modelling}

In contrast to implicit volumetric approaches, 3DGS employs an explicit set of anisotropic Gaussian primitives whose properties are optimized through differentiable rasterization, enabling both direct geometric control and real-time rendering. In surgical environments, EndoGaussian~\cite{liu2024endogaussian} proposes holistic Gaussian initialization and spatio-temporal Gaussian tracking to cope with narrow baselines and rapid soft tissue motion. SurgicalGS~\cite{chen2025surgicalgs} improves geometric accuracy via depth guided initialization and normalized depth regularization to counteract depth compression in inverse depth objectives. Deform3DGS~\cite{yang2024deform3dgs} replaces heavy feature planes with learnable Gaussian bases that parameterize flexible deformations, preserving visual fidelity while cutting training time to approximately one minute per scene. Free-SurGS~\cite{freesurgs} introduces the first SfM-free 3D Gaussian Splatting framework for surgical scene reconstruction, which jointly optimizes camera poses and scene representation by utilizing optical flow as a geometric constraint.

\subsection{SLAM based Method}

NeRF-based SLAM systems such as ENeRF-SLAM~\cite{enerf} and Endo-Depth-and Motion~\cite{Endo-depth-and-motion} adapt implicit fields to online tracking and mapping in endoscopy, but their computational cost limits real-time use. In contrast, EndoGSLAM~\cite{endogslam} leverages the Gaussian representation with differentiable rasterization to maintain frame rates exceeding one hundred frames per second during joint tracking and mapping. EndoFlow-SLAM~\cite{wu2025endoflow} further introduces gaussian flow-constrained optimization, in which the Gaussian map and camera motion are guided by both photometric consistency and optical flow, thereby strengthening temporal alignment and geometric stability in highly deformable sequences.

\section{Methodology}

\subsection{Preliminary}
\subsubsection{3D Gaussian Splatting}
3DGS~\cite{kerbl20233dgs} represents a scene using an explicit collection of anisotropic Gaussian primitives defined in 3D space. Each primitive $G_i$ is parametrized by a mean position $\boldsymbol{\mu}_i \in \mathbb{R}^3$, an opacity $o_i \in [0,1]$, and a covariance matrix $\boldsymbol{\Sigma}_i \in \mathbb{R}^{3 \times 3}$. The spatial density of each Gaussian is given as:
\begin{equation}
  G_i(\mathbf{X}) = o_i \cdot \exp\left\{-\frac{1}{2} (\mathbf{X} - \boldsymbol{\mu}_i)^\top \boldsymbol{\Sigma}_i^{-1} (\mathbf{X} - \boldsymbol{\mu}_i)\right\},
\end{equation}
where $\mathbf{X} \in \mathbb{R}^3$ denotes an arbitrary 3D point. The covariance matrix $\boldsymbol{\Sigma} \in \mathbb{R}^{3 \times 3}$ can be decomposed into a scaling matrix and a rotation quaternion for efficient optimization.

\subsubsection{Wavelet Decomposition}
Wavelet analysis provides a mathematical framework for representing signals through a multi-resolution decomposition. Its fundamental strength lies in achieving a joint time-frequency representation, which allows for the localization of transient features across various scales. In contrast to the Fourier transform, which characterizes a signal's constituent frequencies on a global basis, wavelet transforms employ a family of basis functions called wavelets that are localized in both time and frequency. 


The practical implementation for a discrete image $I \in \mathbb{R}^{H\times W}$ is accomplished via the two-dimensional separable Discrete Wavelet Transform (DWT). This process entails the sequential application of a low-pass filter $h$ and a high-pass filter $g$ along the image's rows and columns. A single level of decomposition has four distinct sub-bands:
\begin{equation}
\begin{aligned}
LL(u,v) &= \sum_{m}\sum_{n} I(m,n)\,h(u{-}m)\,h(v{-}n), \\
LH(u,v) &= \sum_{m}\sum_{n} I(m,n)\,h(u{-}m)\,g(v{-}n), \\
HL(u,v) &= \sum_{m}\sum_{n} I(m,n)\,g(u{-}m)\,h(v{-}n), \\
HH(u,v) &= \sum_{m}\sum_{n} I(m,n)\,g(u{-}m)\,g(v{-}n).
\end{aligned}
\label{eq:wavelet2d}
\end{equation}
The $LL$ sub-band represents a coarse, down-sampled approximation of the original image $I$. In contrast, the set $\{LH, HL, HH\}$ captures high-frequency details corresponding to horizontal, vertical, and diagonal features, respectively. 


\subsection{Spatio-temporal Modeling with 4D Gaussian Splatting}

To capture the complex spatio-temporal dynamics inherent to endoscopic procedures, our work adapts the 4DGS framework~\cite{yang2023real4dgs,lifz2025real}, which models the scene by optimizing a collection of 4D primitives within a unified spacetime volume to represent both spatial structure and temporal evolution simultaneously.

\paragraph{4D primitive and time conditioning.}
We leverage a set of 4D Gaussian primitives to represent the dynamic scene over time\cite{yang2023real4dgs}.
Each primitive is represented by an unnormalized Gaussian density over a spatio-temporal coordinate $(\mathbf{x},t)\in\mathbb{R}^3\times\mathbb{R}$, defined as:

\begin{equation}
p(\mathbf{x},t)
=\exp\!\Big(-\tfrac{1}{2}\big[(\mathbf{x},t)-\boldsymbol{\mu}\big]^{\!\top}
\boldsymbol{\Sigma}^{-1}\big[(\mathbf{x},t)-\boldsymbol{\mu}\big]\Big),
\label{eq:4dgaussian}
\end{equation}
where $\boldsymbol{\mu}=(\boldsymbol{\mu}_{x},\mu_{t})$ is the 4D mean, and $\boldsymbol{\Sigma}\in\mathbb{R}^{4\times 4}$ is 4D covariance . We partition $\boldsymbol{\Sigma}$ into spatial, cross, and temporal blocks as
$\boldsymbol{\Sigma}_{x,x}\in\mathbb{R}^{3\times 3}$, $\boldsymbol{\Sigma}_{x,t}\in\mathbb{R}^{3\times 1}$, and $\boldsymbol{\Sigma}_{t,t}\in\mathbb{R}$.

\paragraph{Time-Evolved Appearance Spherindrical Harmonics}
In 4DGS, view-dependent appearance is expanded in 4D spherindrical harmonics that couple spherical harmonics over viewing direction with a cosine-based temporal component, which implicitly fixes each temporal atom to a zero-phase origin. Building on this formulation, we introduce Time-Evolved Appearance Spherindrical Harmonics(TEASH), which incorporates a learnable phase for each temporal frequency. This approach produces a phase-adaptive Fourier atom while preserving the original spatial structure of the SH. Consequently, 4D TEASH $Z_{nl}^{m}(t,\theta,\phi)$ can be expressed as:
 
\begin{equation}
Z_{nl}^{m}(t,\theta,\phi)\;=\;Y_{l}^{m}(\theta,\phi)\,\cos\!\big(\omega_{n}\,t+\varphi_{n}\big),
\end{equation}
where $Y_{l}^{m}$ denotes the real spherical harmonic for direction $(\theta,\phi)$, $\omega_{n}=2\pi n/T$ is the $n$-th temporal angular frequency over period $T$, and $\varphi_{n}$ is a learnable phase. This parameterization aligns each temporal component with the observed phase of the signal, since 
$\cos(\omega t+\varphi)$
 spans the sine–cosine pair at frequency $\omega$ without introducing extra coefficients. In practice, TEASH retains the orthogonal angular basis while improving temporal fit to generate a concise and phase-consistent appearance model for dynamic content.

\subsection{Flow induced geometric constraint}

\paragraph{3D Scene Flow and Estimated 2D Optical Flow}

Our 4D Gaussian Splatting representation models a dynamic scene by defining the trajectory of each Gaussian primitive over time. For any given Gaussian $i$, its 3D center position is a function of time, denoted as $\mathbf{P}_i(t) \in \mathbb{R}^3$ according to Eq.~\eqref{eq:4dgaussian}. The instantaneous 3D velocity field of the scene is known as scene flow. We define the discrete scene flow vector $\mathbf{S}_i$ for Gaussian $i$ between two time steps, $t_1$ and $t_2$, as the difference in its 3D position:
\begin{equation}
    \mathbf{S}_i = \mathbf{P}_i(t_2) - \mathbf{P}_i(t_1).
\end{equation}

To obtain a 2D optical flow representation from this 3D scene flow, we project the 3D Gaussian centers at both time steps onto their respective 2D image planes. Let $\Pi(\cdot)$ be the projection function that maps a 3D world point to 2D pixel coordinates, using the camera's intrinsic matrix $\mathbf{K}$ and extrinsic matrix $\mathbf{E}$. The 2D projection of Gaussian $i$ at time $t$ is
\begin{equation}
\mathbf{p}_i(t)\;=\;\Pi\!\big(\mathbf{P}_i(t),\,\mathbf{K}_t,\,\mathbf{E}_t\big).
\end{equation}

The estimated 2D optical flow vector $\mathbf{f}_{est, i}$ for an individual Gaussian is the displacement of its projected center:
\begin{equation}
    \mathbf{f}_{est, i} = \mathbf{p}_i(t_2) - \mathbf{p}_i(t_1).
\end{equation}

For a 2D pixel $\mathbf{p}$, the per-pixel flow is obtained by standard front-to-back alpha compositing:
\begin{equation}
\hat{\mathbf{f}}(\mathbf{p}) \;=\; \mathbf{F}_{\text{est}}(\mathbf{p})
\;=\; \sum_{i\in\mathcal{G}(\mathbf{p})} w_i(\mathbf{p})\,\mathbf{f}_{\text{est},i},
\end{equation}
where $\mathcal{G}(\mathbf{p})$ denotes the set of Gaussians contributing to pixel $\mathbf{p}$, sorted by the visibility ordering.

\paragraph{Ground Truth Optical Flow Supervision}

To supervise the estimated dense flow $\hat{\mathbf{f}}(\mathbf{p})$, we generate ground-truth optical flow. Specifically, we first render the scene at times $t_1$ and $t_2$ with the 4D Gaussian Splatting pipeline to generate two consecutive frames $\mathbf{I}(t_1)$ and $\mathbf{I}(t_2)$. We use the off-the-shelf method GMFlow to get optical flow ${\mathbf{f}}(\mathbf{p})$ as pseudo GT.

To enhance robustness, we filter unreliable pixel correspondences using a bi-directional consistency check on the pseudo ground-truth optical flow between frames $\mathbf{I}(t_1)$ and $\mathbf{I}(t_2)$. This process generates a binary validity mask, $\mathbf{M}_{valid}$, to identify pixels with consistent flow. The final flow loss is subsequently computed only on the regions validated by this $\mathbf{M}_{valid}$ mask:

\begin{equation}
  \mathcal{L}_{\text{flow}}
  = \left\| \mathbf{M}_{valid} \odot \hat{\mathbf{f}}(\mathbf{p})
   - \mathbf{M}_{valid} \odot \mathbf{f}(\mathbf{p}) \right\|_2 .
\end{equation}

\begin{figure*}[!h]
    \centering
    \includegraphics[width=0.85\textwidth]{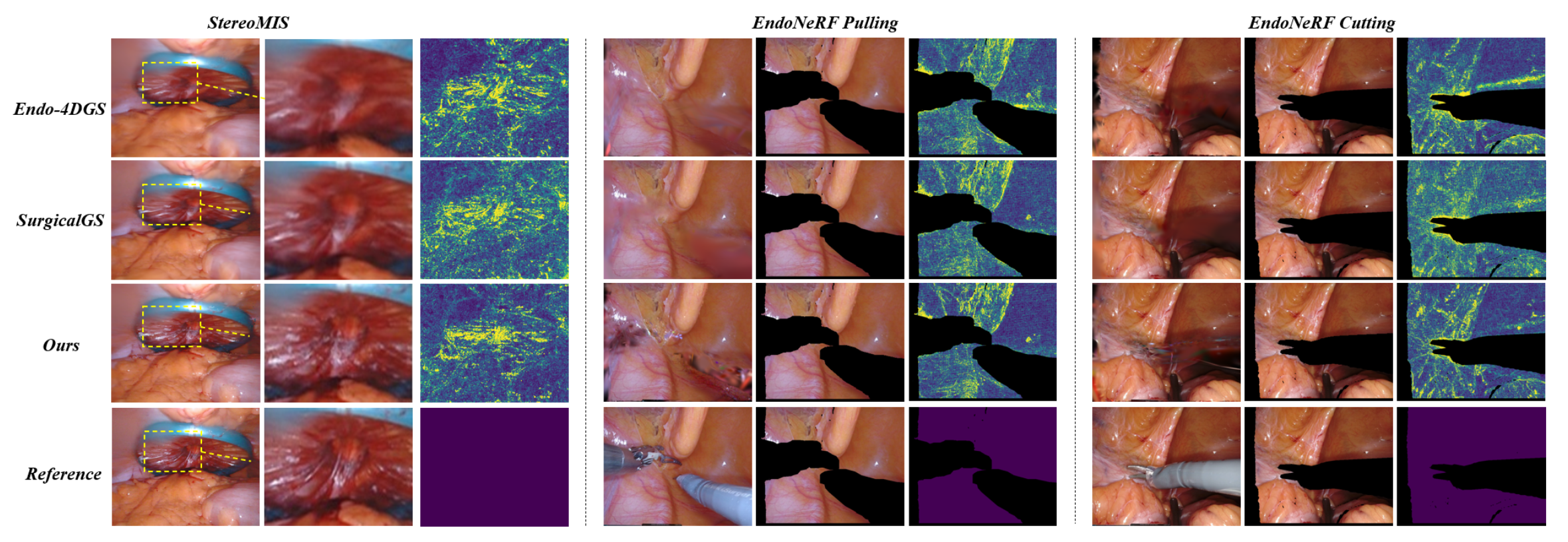}
    \caption{\textbf{Qualitative results for novel view synthesis.} Left: Results on the StereoMIS~\cite{stereomis} dataset, with magnified details. Center and Right: Comparison of the Cutting and Pulling sequences from the EndoNeRF~\cite{endonerf} dataset, respectively. The last column of each sequence is the error map, dark purple indicates low error, and yellow indicates high error.}
    \label{fig:visEndoNerf}
\end{figure*}

\subsection{Rational Wavelet Decomposition}

\subsubsection{Limitations of traditional dyadic wavelets.}

Dyadic wavelet systems (orthogonal or biorthogonal) partition scale in octaves $2^{j}$, producing a regular logarithmic tiling amenable to fast filter banks. For endoscopic images—where narrow specular spikes overlay low-frequency albedo—these octave steps from $2^{j}$ to $2^{j+1}$ are often too coarse to isolate sharp peaks while preserving smooth content. Linear-phase biorthogonal variants do not resolve this, since their scale quantization is still dyadic; hence, dyadic analysis can be ill-suited when key features lie between octaves.

To address this limitation, a \emph{rational wavelet} transform offers a more flexible multiresolution analysis by employing a rational dilation factor, allowing for a denser tiling of the time-frequency plane.

\paragraph{Rational Scaling Principle}
The core of the rational wavelet transform is the rational scale factor $a$, defined as:
\begin{equation}
    a = \frac{p}{q} = \frac{q+1}{q}, \qquad q \in \mathbb{N}, q \ge 1
\end{equation}
where $p$ and $q$ are coprime integers. This construction generates a sequence of scales $a^j$ that are more closely spaced than their dyadic counterparts.

\paragraph{Time-Domain Filter Design}

Formal rational wavelet constructions, such as those based on the Meyer wavelet, are defined by their complex spectra in the frequency domain. Consequently, their corresponding time-domain filter coefficients lack a simple analytical form and must be computed numerically via an Inverse Fourier Transform. For a computationally efficient implementation, particularly within a deep learning loss function, we adopt a pragmatic approach in the time domain. We design a set of Meyer-like filters directly in the discrete time domain. These filters are inspired by the formal wavelet analysis~\cite{yu2018rational} but are constructed from Gaussian functions for numerical stability and straightforward implementation. The filter design is parameterized by the rational scale factor $a$, from which we define a scale-adaptive bandwidth parameter $\sigma = 1/a$. The filters are defined over a discrete time index $t$.

\noindent \textbf{Low-Pass Scaling Filter ($h_0$):} A smooth, zero-phase low-pass filter, analogous to a scaling function filter, is constructed from a normalized Gaussian function. This filter $h_0$ captures the low-frequency approximation of the signal.
\begin{equation}
    h_0(t) = \frac{\exp\left(-\frac{1}{2}\left(\frac{t}{\sigma}\right)^2\right)}{\sum_{u} \exp\left(-\frac{1}{2}\left(\frac{u}{\sigma}\right)^2\right)}.
    \label{eq:h0}
\end{equation}

\vspace{-5pt}

\noindent\textbf{High-Pass Wavelet Filters ($h_1, g$):} Two distinct filters are defined to capture signal details at different orientations, similar to wavelet function filters.
The first, $h_1(t)$, is a derivative-of-Gaussian (DoG) type filter, which is zero-mean and effective at emphasizing edges.
\begin{equation}
    h_1(t) = \frac{-t \cdot \exp\left(-\frac{1}{2}\left(\frac{t}{\sigma}\right)^2\right)}{\sum_{u} \left| -u \cdot \exp\left(-\frac{1}{2}\left(\frac{u}{\sigma}\right)^2\right) \right|}.
    \label{eq:h1}
\end{equation}
The second, $g(t)$, is a smooth, Meyer-like band-pass filter constructed as a modulated Gaussian. Its center frequency depends on the rational factor $a$.
\begin{equation}
    g(t) = \frac{\sin(\pi t / a) \cdot \exp\left(-\frac{1}{2}\left(\frac{t}{2\sigma}\right)^2\right)}{\sum_{u} \left| \sin(\pi u / a) \cdot \exp\left(-\frac{1}{2}\left(\frac{u}{2\sigma}\right)^2\right) \right|}.
    \label{eq:g}
\end{equation}
In the above definitions, $u$ is the summation over all the discrete sample points of the filter kernel.

\begin{figure*}[!t]
    \centering
    \includegraphics[width=0.75\textwidth]{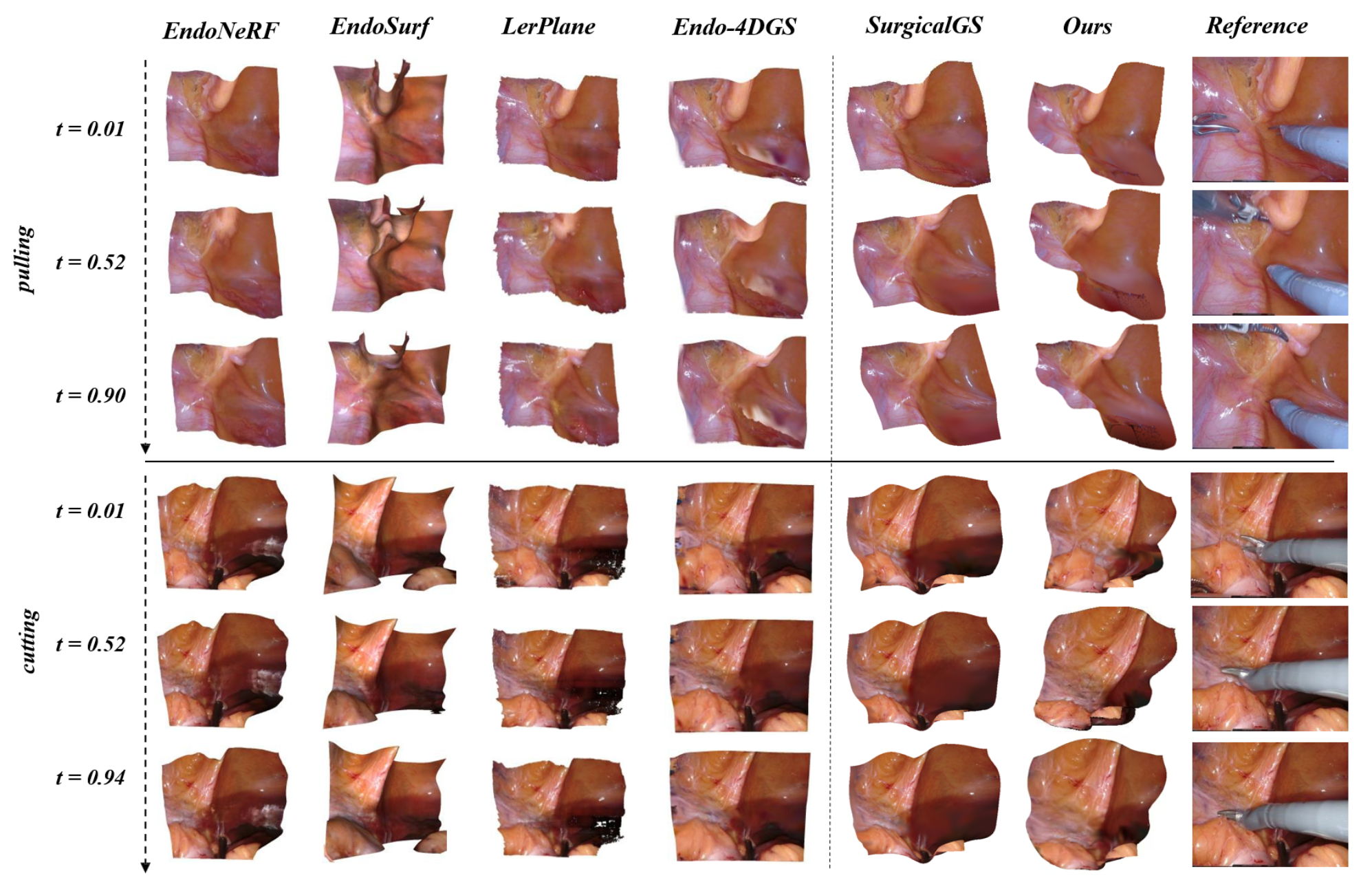}
    \caption{\textbf{Qualitative 3D Model comparison on the EndoNeRF dataset}}
    \label{fig:pipeline}
\end{figure*}
\paragraph{Separable 2D Transform}
For a 2D signal image $I$, a separable wavelet transform is constructed from the tensor product of 1D analysis filters applied sequentially to the signal's rows and columns. This decomposition is performed iteratively over $J$ levels, indexed by $j \in \{0, 1, \dots, J-1\}$, to produce a multi-level signal representation.

The input to the transform at level $j$ is denoted by $I^{(j)}$, with the original signal serving as the base level, $I^{(0)} = I$. At each level of analysis, the input $I^{(j)}$ is decomposed into four distinct frequency sub-bands. This decomposition is achieved by first performing 1D convolutions along the rows ($*_{r}$) and columns ($*_{c}$), followed by a rational downsampling. 
The sub-bands are computed according to Eq.~\eqref{eq:wavelet2d} as follow:

\begin{equation}
\begin{aligned}
    LL^{(j)} \,&=\, \big((I^{(j)} *_{r} h_0) *_{c} h_0\big) \\
    LH^{(j)} \,&=\, \big((I^{(j)} *_{r} h_0) *_{c} h_1\big) \\
    HL^{(j)} \,&=\, \big((I^{(j)} *_{r} h_1) *_{c} h_0\big) \\
    HH^{(j)} \,&=\, \big((I^{(j)} *_{r} g)\; *_{c} g\big)
\end{aligned}
\label{eq:wavelet}
\end{equation}

The approximation sub-band $LL^{(j)}$ is propagated to the next level to serve as the input for the subsequent decomposition,  $I^{(j+1)} = LL^{(j)}$. 

\paragraph{Wavelet-domain loss.}

We utilize the rational orthogonal wavelet transform with two-level decomposition. We utilize $L_2$ loss based on the frequency map from the rendered image and ground truth image. The discrepancies of each frequency between rendered image \(\hat{I}\) and ground truth \(I\) can be expressed as: 
\begin{equation}
    \mathcal{L}_\text{wavelet} = 
\sum_{x \in \{LL,LH, HL, HH\}} \lambda_x \left\| W_x(I_t) - W_x\left(\hat{I}_t\right) \right\|,
\end{equation}
where \(W_x\) represents the wavelet transformation that extract component \(x\) from an image and the $\lambda_x$ denotes the weight for each wavelet component.

\subsection{Overall Training Objective function}
\label{subsec:pipeline}
With all defined loss terms, the overall training objective can be formulated as follows:
\begin{equation}
\mathcal{L} \;=\; 
\lambda_{\text{rgb}} \,\mathcal{L}_{\text{rgb}}
+ \lambda_{\text{depth}} \,\mathcal{L}_{\text{depth}}
+ \lambda_{\text{flow}} \,\mathcal{L}_{\text{flow}}
+ \lambda_{\text{wavelet}} \,\mathcal{L}_{\text{wavelet}} ,
\label{eq:total_loss}
\end{equation}
where $\mathcal{L}_{\text{rgb}}$ and $\mathcal{L}_{\text{depth}}$ denote the original losses in 3DGS~\cite{kerbl20233dgs}, while $\mathcal{L}_{\text{flow}}$ and $\mathcal{L}_{\text{wavelet}}$ correspond to the proposed flow constraint and wavelet constraint, respectively.

\vspace{-5pt}
\begin{table*}[!t]
\centering
\caption{Quantitative results on the EndoNeRF dataset~\cite{endonerf}. Best results are in \textbf{bold}.}
\label{tab:endonerf}
\begin{tabular}{@{}l ccc ccc c@{}}
\toprule
\multirow{2}{*}{Models} & \multicolumn{3}{c}{EndoNeRF-Cutting} & \multicolumn{3}{c}{EndoNeRF-Pulling} & \multirow{2}{*}{FPS $\uparrow$} \\
\cmidrule(lr){2-4} \cmidrule(lr){5-7}
& PSNR $\uparrow$ & SSIM $\uparrow$ & LPIPS $\downarrow$ & PSNR $\uparrow$ & SSIM $\uparrow$ & LPIPS $\downarrow$ & \\
\midrule
EndoNeRF~\cite{endonerf}      & 35.84 & 0.942 & 0.057  & 35.43 & 0.939 & 0.064  & 0.2 \\
EndoSurf~\cite{zha2023endosurf}     & 34.89 & 0.952 & 0.107  & 34.91 & 0.955 & 0.120  & 0.04 \\
LerPlane-32k~\cite{yang2023neural}  & 34.66 & 0.923 & 0.071  & 31.77 & 0.910 & 0.071  & 1.5 \\
Endo-4DGS~\cite{huang2024endo}                  & 36.56 & 0.955 & 0.032  & 37.85 & 0.959 & 0.043  & 100 \\
EndoGS~\cite{zhu2024endogs}                               & 37.16 & 0.953 & 0.045  & 36.19 & 0.941 & 0.041  & 70 \\
SurgicalGS~\cite{chen2025surgicalgs}                  & 38.31 & 0.962 & 0.062 & 38.05 & 0.959 & 0.062 & \textbf{194} \\
\midrule
\textbf{Ours} & \textbf{38.93} & \textbf{0.981} & \textbf{0.010} & \textbf{38.51} & \textbf{0.969} & \textbf{0.021} & 86 \\
\bottomrule
\end{tabular}
\end{table*}

\section{Experiments}
\subsection{Implementation Details}
\subsubsection{Experimental Setup}
All experiments were conducted on a single NVIDIA RTX 4090 GPU. We employed the Adam optimizer~\cite{adam} for optimizing Gaussian primitives. For the attributes of the Gaussians, we follow the hyperparameter settings from Kerbl et al.~\cite{kerbl20233dgs}, such as the learning rates and the threshold for density control. To ensure a fair comparison, the training iteration strategy was kept consistent with~\cite{yang2023real4dgs}. To validate the generalizability of our method, we applied the fixed set of hyperparameters, along with a consistent loss function and initialization strategy, across all datasets without any per-scene tuning for a fixed number of iterations.

\subsubsection{Datasets}
We quantitatively and qualitatively evaluate our proposed method on two publicly available datasets for dynamic endoscopic scene reconstruction: EndoNeRF and StereoMIS. (1) The EndoNeRF dataset~\cite{endonerf} consists of stereo endoscopic sequences from two human prostatectomy procedures. Captured from a stationary viewpoint, this dataset provides challenging scenarios involving significant non-rigid tissue deformation and frequent tool occlusions. Each sequence is supplied with estimated depth maps, which are pre-calculated using stereo matching algorithms, and manually annotated masks to segment surgical tools from the scene.
(2) The StereoMIS dataset~\cite{stereomis} features eleven stereo video sequences from live porcine surgeries, recorded with the da Vinci Xi surgical system. This dataset is particularly characterized by large-scale and complex tissue deformations throughout the procedures. Similar to EndoNeRF, the dataset is annotated with corresponding tool segmentation masks. For our experiments, we extract the continuous 800 to 1000 frames from the first scene following ~\cite{huang2024endo}. For all scenes from the EndoNeRF and StereoMIS datasets, the frames were partitioned into training and testing sets following the 7:1 ratio.

\subsubsection{Evaluation Metrics}
We evaluate the quality of our novel view synthesis using three image quality metrics: Peak Signal-to-Noise Ratio (PSNR), Learned Perceptual Image Patch Similarity (LPIPS), and Structural Similarity Index Measure (SSIM).
In addition to accuracy metrics, we also report practical performance indicators, including inference speed measured in Frames Per Second (FPS).

\vspace{-0.6em}

\subsection{Quantitative and qualitative results}

We evaluated our method, EndoWave, against six representative reconstructing surgical scenes methods on the EndoNeRF and StereoMIS datasets. The compared methods include NeRF-based approaches and Gaussian Splatting relevant frameworks. The performance metrics, including PSNR, SSIM, LPIPS, and rendering speed (FPS), are detailed in Table \ref{tab:endonerf} and Table \ref{tab:StereoMIS}.

\begin{table}[htb]
\centering
\caption{Quantitative results on the StereoMIS dataset~\cite{stereomis}. Best results are in \textbf{bold}.}
\vspace{-0.5em}
\label{tab:StereoMIS}
\begin{tabular}{@{}l ccc c@{}}
\toprule
Models & PSNR $\uparrow$ & SSIM $\uparrow$ & LPIPS $\downarrow$ & FPS $\uparrow$ \\
\midrule
EndoNeRF~\cite{endonerf}      & 21.49   & 0.622   & 0.360  & 0.2 \\
EndoSurf~\cite{zha2023endosurf}     & 29.87   & 0.809   & 0.303  & 0.04 \\
LerPlane-32k~\cite{yang2023neural}  & 30.80   & 0.826   & 0.174  & 1.7 \\
Endo-4DGS~\cite{huang2024endo}                           & 32.69   & 0.850   & 0.148  & 100 \\
SurgicalGS~\cite{chen2025surgicalgs}                          & 31.60   & 0.854   & 0.263  & \textbf{183} \\
\midrule
\textbf{Ours}              & \textbf{33.26} & \textbf{0.9126} & \textbf{0.075} & 77 \\
\bottomrule
\end{tabular}
\end{table}

\vspace{-2pt}
On the EndoNeRF dataset~\cite{endonerf}, our method demonstrates superior reconstruction quality while operating at a real-time frame rate of 86 FPS. In contrast, existing methods, such as EndoNeRF~\cite{endonerf} and EndoSurf~\cite{zha2023endosurf}, are limited to 0.2 and 0.04 FPS, respectively, which hinders their practical deployment. Leveraging the efficient scene representation of 3D Gaussian Splatting (3DGS), our approach not only preserves this computational advantage but also achieves high-fidelity novel view synthesis. Specifically, on the Cutting and Pulling sequences, our method attains PSNR scores of 38.93 and 38.51, respectively. As illustrated in Figure~\ref{fig:visEndoNerf}, we provide ground-truth-masked error maps where dark purple and yellow represent low and high reconstruction error, respectively.
To further validate the generalization and robustness of our approach, we conducted evaluations on the StereoMIS dataset~\cite{stereomis}. As reported in Table~\ref{tab:StereoMIS}, our method achieves state-of-the-art performance, with a PSNR of 33.26, an SSIM of 0.9126, and an LPIPS of 0.0554. For qualitative analysis, we present magnified views of regions with significant discrepancies in the error maps, as illustrated in the left of Figure~\ref{fig:visEndoNerf}.

This significant performance improvement is a direct result of our core contributions. Firstly, the direct time-conditioned 4DGS representation more accurately models complex non-rigid tissue deformations compared to approaches that rely on a canonical space and a deformation field. Secondly, the integration of an optical flow-based geometric constraint ensures temporal coherence across frames. Finally, the lower LPIPS score is attributed to our proposed rational orthogonal wavelets, which effectively separate the high and low-frequency components of endoscopic scenes.

\vspace{-5pt}

\subsection{Abaltion Study}

We conducted an ablation study on the Pulling sequence of the EndoNeRF dataset to further analyze the contribution of our design. Starting from the full model, we progressively removed the proposed components and loss terms and performed extensive ablations. In addition to disabling our wavelet constraint $\mathcal{L}_{\text{wavelet}}$ and flow constraint $\mathcal{L}_{\text{flow}}$, we also examined the effect of removing the depth constraint inherited from the original framework. As shown in Table~\ref{tab:ablation}, excluding any single component consistently degrades performance, highlighting the critical role each plays in improving reconstruction quality, accuracy, and overall robustness.

\begin{table}[htb]
\centering
\caption{Ablation experiments of the proposed method on EndoNeRF dataset~\cite{endonerf}. The best results are in bold.}
\vspace{-1em}
\label{tab:ablation}
\resizebox{\columnwidth}{!}{%
\begin{tabular}{c|c|c|ccc}
\toprule
\multirow{2}{*}{\makecell[c]{Flow \\Constraint}} &
\multirow{2}{*}{\makecell[c]{Wavelet\\Constraint}} &
\multirow{2}{*}{\makecell[c]{Depth\\Constraint}} &
\multicolumn{3}{c}{EndoNeRF-Pulling} \\ \cline{4-6}
& & & PSNR $\uparrow$ & SSIM $\uparrow$ & LPIPS $\downarrow$ \\
\midrule
\XSolidBrush & \XSolidBrush & \Checkmark & 37.44 & 0.948 & 0.027 \\
\Checkmark   & \XSolidBrush & \XSolidBrush & 37.25 & 0.948 & 0.026 \\
\XSolidBrush & \Checkmark   & \XSolidBrush & 36.93 & 0.946 & 0.030 \\
\Checkmark   & \XSolidBrush & \Checkmark   & 37.47 & 0.947 & 0.029 \\
\XSolidBrush & \Checkmark   & \Checkmark   & 37.20 & 0.946 & 0.029 \\
\Checkmark   & \Checkmark   & \XSolidBrush & 37.56 & 0.950 & 0.027 \\
\Checkmark   & \Checkmark   & \Checkmark   & \textbf{38.51} & \textbf{0.969} & \textbf{0.021} \\ 
\bottomrule
\end{tabular}%
}
\end{table}

\vspace{-1em}

\begin{table}[htb]
\centering
\caption{Effect of wavelet constraint on StereoMIS dataset (second sequence) with Endo-4DGS~\cite{huang2024endo}. Best results are in \textbf{bold}.}
\vspace{-0.7em}
\label{tab:endo4dgs_ablation}
\setlength{\tabcolsep}{5pt}
\renewcommand{\arraystretch}{1.05}
\begin{tabular}{@{}l c c c@{}}
\toprule
Method & PSNR $\uparrow$ & SSIM $\uparrow$ & LPIPS $\downarrow$ \\
\midrule
Endo-4DGS~\cite{huang2024endo} & 31.49 & 0.837 & 0.211 \\
Endo-4DGS + $\mathcal{L}_{\text{wavelet}}$ & \textbf{31.72} & \textbf{0.839} & \textbf{0.203} \\
\bottomrule
\end{tabular}
\end{table}

\vspace{1pt}
We further evaluated generalization on the StereoMIS~\cite{stereomis} dataset by incorporating the proposed wavelet constraint directly into Endo-4DGS~\cite{huang2024endo}. The results Table~\ref{tab:endo4dgs_ablation} show effective performance when only $\mathcal{L}_{\text{wavelet}}$ is integrated, without modifying other parts.

\vspace{-1em}
\section{Conclusion}

In this paper, we introduced EndoWave, a spatio-temporal Gaussian Splatting method for endoscopic reconstruction. Operating directly within a 4D domain, our approach enhances standard photometric training by incorporating supervision sensitive to both motion and frequency. To achieve this, the method estimates per-primitive scene flow from temporal evolution, projects this flow into the image space, and enforces consistency with externally computed optical flow. Furthermore, a two-level rational orthogonal wavelet loss constrains both low-frequency global appearance and high-frequency details, effectively mitigating artifacts from sources like specular reflections. Quantitative evaluations on the EndoNeRF and StereoMIS datasets show the effectiveness of our approach. EndoWave achieves a PSNR of 38.93 dB on the EndoNeRF Cutting sequence and 38.51 dB on the Pulling sequence, with interactive rendering at 86 FPS. On StereoMIS, it obtains 33.26 dB PSNR and 0.9126 SSIM at 77 FPS. Further analysis through ablation studies validates that the optical flow guidance, wavelet supervision, and depth regularization each provide complementary performance benefits.

This work relies on pseudo ground-truth optical flow and a fixed two-level wavelet configuration. Future work will investigate self-supervised motion cues that reduce dependence on external estimators, as well as adaptive scale selection in rational wavelets, including robustness under heavy instrument occlusion and rapid camera motions. We also plan to explore appearance models with richer temporal bases and to validate the method on larger multi-institutional datasets to assess generalization.

\vspace{-1em}

\section{Acknowledgments}
This research was partially funded by the Postgraduate Research Scholarship (PGRS) at Xi’an Jiaotong-Liverpool University, contract number TW5A2312001 and FOS2104JP08. This research was partially funded by Suzhou Municipal Key Laboratory Broadband Wireless Access Technology (BWAT) and JITRI Supervision Support Fund (JSF10120220008) of XJTLU-JITRI Academy.

\vspace{-0.5em}
\bibliographystyle{IEEEtran}
\bibliography{bibmendo}

@inproceedings{xu2024review,
  title={A review of 3d reconstruction techniques for deformable tissues in robotic surgery},
  author={Xu, Mengya and Guo, Ziqi and Wang, An and Bai, Long and Ren, Hongliang},
  booktitle={International Conference on Medical Image Computing and Computer-Assisted Intervention},
  pages={157--167},
  year={2024},
  organization={Springer}
}

@inproceedings{lifz2025real,
  title={Real-Time Spatio-Temporal Reconstruction of Dynamic Endoscopic Scenes with 4D Gaussian Splatting},
  author={Li, Fengze and He, Jishuai and Ma, Jieming and Wu, Zhijing},
  booktitle={2025 IEEE 22nd International Symposium on Biomedical Imaging (ISBI)},
  pages={1--5},
  year={2025},
  organization={IEEE}
}

@inproceedings{huang2024endo,
  title={Endo-4dgs: Endoscopic monocular scene reconstruction with 4d gaussian splatting},
  author={Huang, Yiming and Cui, Beilei and Bai, Long and Guo, Ziqi and Xu, Mengya and Islam, Mobarakol and Ren, Hongliang},
  booktitle={International Conference on Medical Image Computing and Computer-Assisted Intervention},
  pages={197--207},
  year={2024},
  organization={Springer}
}

@inproceedings{wang2024endogslam,
  title={Endogslam: Real-time dense reconstruction and tracking in endoscopic surgeries using gaussian splatting},
  author={Wang, Kailing and Yang, Chen and Wang, Yuehao and Li, Sikuang and Wang, Yan and Dou, Qi and Yang, Xiaokang and Shen, Wei},
  booktitle={International Conference on Medical Image Computing and Computer-Assisted Intervention},
  pages={219--229},
  year={2024},
  organization={Springer}
}

@article{mildenhall2021nerf,
  title={Nerf: Representing scenes as neural radiance fields for view synthesis},
  author={Mildenhall, Ben and Srinivasan, Pratul P and Tancik, Matthew and Barron, Jonathan T and Ramamoorthi, Ravi and Ng, Ren},
  journal={Communications of the ACM},
  volume={65},
  number={1},
  pages={99--106},
  year={2021},
  publisher={ACM New York, NY, USA}
}

@article{yu2018rational,
  title={Rational-orthogonal-wavelet-based active sonar pulse and detector design},
  author={Yu, Limin and Ma, Fei and Lim, Enggee and Cheng, En and White, Langford B},
  journal={IEEE Journal of Oceanic Engineering},
  volume={44},
  number={1},
  pages={167--178},
  year={2018},
  publisher={IEEE}
}

@article{kerbl20233dgs,
  title={3d gaussian splatting for real-time radiance field rendering.},
  author={Kerbl, Bernhard and Kopanas, Georgios and Leimk{\"u}hler, Thomas and Drettakis, George},
  journal={ACM Trans. Graph.},
  volume={42},
  number={4},
  pages={139--1},
  year={2023}
}

@inproceedings{endonerf,
  title={Neural rendering for stereo 3d reconstruction of deformable tissues in robotic surgery},
  author={Wang, Yuehao and Long, Yonghao and Fan, Siu Hin and Dou, Qi},
  booktitle={MICCAI},
  pages={431--441},
  year={2022},
  organization={Springer}
}

@article{Endo-depth-and-motion,
  title={Endo-depth-and-motion: Reconstruction and tracking in endoscopic videos using depth networks and photometric constraints},
  author={Recasens, David and Lamarca, Jos{\'e} and F{\'a}cil, Jos{\'e} M and Montiel, JMM and Civera, Javier},
  journal={IEEE Robotics and Automation Letters},
  volume={6},
  number={4},
  pages={7225--7232},
  year={2021},
  publisher={IEEE}
}

@inproceedings{endogslam,
  title={Endogslam: Real-time dense reconstruction and tracking in endoscopic surgeries using gaussian splatting},
  author={Wang, Kailing and Yang, Chen and Wang, Yuehao and Li, Sikuang and Wang, Yan and Dou, Qi and Yang, Xiaokang and Shen, Wei},
  booktitle={International Conference on Medical Image Computing and Computer-Assisted Intervention},
  pages={219--229},
  year={2024},
  organization={Springer}
}

@article{stereomis,
  title={Learning how to robustly estimate camera pose in endoscopic videos},
  author={Hayoz, Michel and Hahne, Christopher and Gallardo, Mathias and Candinas, Daniel and Kurmann, Thomas and Allan, Maximilian and Sznitman, Raphael},
  journal={International journal of computer assisted radiology and surgery},
  volume={18},
  number={7},
  pages={1185--1192},
  year={2023},
  publisher={Springer}
}

@article{adam,
  title={Adam: A method for stochastic optimization},
  author={Kingma, Diederik P},
  journal={arXiv preprint arXiv:1412.6980},
  year={2014}
}

@article{ulsr,
  title={ULSR-GS: Ultra Large-scale Surface Reconstruction Gaussian Splatting with Multi-View Geometric Consistency},
  author={Li, Zhuoxiao and Yao, Shanliang and Wu, Taoyu and Yue, Yong and Zhao, Wufan and Qin, Rongjun and Garcia-Fernandez, Angel F and Levers, Andrew and Zhu, Xiaohui},
  journal={arXiv preprint arXiv:2412.01402},
  year={2024}
}

@inproceedings{freesurgs,
  title={Free-SurGS: SfM-Free 3D Gaussian Splatting for Surgical Scene Reconstruction},
  author={Guo, Jiaxin and Wang, Jiangliu and Kang, Di and Dong, Wenzhen and Wang, Wenting and Liu, Yun-hui},
  booktitle={International Conference on Medical Image Computing and Computer-Assisted Intervention},
  pages={350--360},
  year={2024},
  organization={Springer}
}

@article{enerf,
  title={ENeRF-SLAM: a dense endoscopic slam with neural implicit representation},
  author={Shan, Jiwei and Li, Yirui and Xie, Ting and Wang, Hesheng},
  journal={IEEE Transactions on Medical Robotics and Bionics},
  year={2024},
  publisher={IEEE}
}

@article{liu2024endogaussian,
  title={Endogaussian: Real-time gaussian splatting for dynamic endoscopic scene reconstruction},
  author={Liu, Yifan and Li, Chenxin and Yang, Chen and Yuan, Yixuan},
  journal={arXiv preprint arXiv:2401.12561},
  year={2024}
}

@inproceedings{wu4dgs,
  title={4d gaussian splatting for real-time dynamic scene rendering},
  author={Wu, Guanjun and Yi, Taoran and Fang, Jiemin and Xie, Lingxi and Zhang, Xiaopeng and Wei, Wei and Liu, Wenyu and Tian, Qi and Wang, Xinggang},
  booktitle={Proceedings of the IEEE/CVF conference on computer vision and pattern recognition},
  pages={20310--20320},
  year={2024}
}

@inproceedings{yang2024deform3dgs,
  title={Deform3dgs: Flexible deformation for fast surgical scene reconstruction with gaussian splatting},
  author={Yang, Shuojue and Li, Qian and Shen, Daiyun and Gong, Bingchen and Dou, Qi and Jin, Yueming},
  booktitle={International Conference on Medical Image Computing and Computer-Assisted Intervention},
  pages={132--142},
  year={2024},
  organization={Springer}
}

@inproceedings{chen2025surgicalgs,
  title={Surgicalgs: Dynamic 3d gaussian splatting for accurate robotic-assisted surgical scene reconstruction},
  author={Chen, Jialei and Zhang, Xin and Hoque, Mobarak I and Vasconcelos, Francisco and Stoyanov, Danail and Elson, Daniel S and Huang, Baoru},
  booktitle={International Conference on Medical Image Computing and Computer-Assisted Intervention},
  pages={572--582},
  year={2025},
  organization={Springer}
}

@inproceedings{long2021dssr,
  title={E-dssr: efficient dynamic surgical scene reconstruction with transformer-based stereoscopic depth perception},
  author={Long, Yonghao and Li, Zhaoshuo and Yee, Chi Hang and Ng, Chi Fai and Taylor, Russell H and Unberath, Mathias and Dou, Qi},
  booktitle={International Conference on Medical Image Computing and Computer-Assisted Intervention},
  pages={415--425},
  year={2021},
  organization={Springer}
}

@inproceedings{zha2023endosurf,
  title={Endosurf: Neural surface reconstruction of deformable tissues with stereo endoscope videos},
  author={Zha, Ruyi and Cheng, Xuelian and Li, Hongdong and Harandi, Mehrtash and Ge, Zongyuan},
  booktitle={International conference on medical image computing and computer-assisted intervention},
  pages={13--23},
  year={2023},
  organization={Springer}
}

@article{guo2024uc,
  title={UC-NeRF: Uncertainty-aware Conditional Neural Radiance Fields from Endoscopic Sparse Views},
  author={Guo, Jiaxin and Wang, Jiangliu and Wei, Ruofeng and Kang, Di and Dou, Qi and Liu, Yun-hui},
  journal={IEEE Transactions on Medical Imaging},
  year={2024},
  publisher={IEEE}
}

@inproceedings{yang2023neural,
  title={Neural lerplane representations for fast 4d reconstruction of deformable tissues},
  author={Yang, Chen and Wang, Kailing and Wang, Yuehao and Yang, Xiaokang and Shen, Wei},
  booktitle={International Conference on Medical Image Computing and Computer-Assisted Intervention},
  pages={46--56},
  year={2023},
  organization={Springer}
}

@article{yang2023real4dgs,
  title={Real-time photorealistic dynamic scene representation and rendering with 4d gaussian splatting},
  author={Yang, Zeyu and Yang, Hongye and Pan, Zijie and Zhang, Li},
  journal={arXiv preprint arXiv:2310.10642},
  year={2023}
}

@article{wu2025endoflow,
  title={EndoFlow-SLAM: Real-Time Endoscopic SLAM with Flow-Constrained Gaussian Splatting},
  author={Wu, Taoyu and Miao, Yiyi and Li, Zhuoxiao and Zhao, Haocheng and Dang, Kang and Su, Jionglong and Yu, Limin and Li, Haoang},
  journal={arXiv preprint arXiv:2506.21420},
  year={2025}
}

@inproceedings{zhu2024endogs,
  title={Endogs: Deformable endoscopic tissues reconstruction with gaussian splatting},
  author={Zhu, Lingting and Wang, Zhao and Cui, Jiahao and Jin, Zhenchao and Lin, Guying and Yu, Lequan},
  booktitle={International Conference on Medical Image Computing and Computer-Assisted Intervention},
  pages={135--145},
  year={2024},
  organization={Springer}
}

@inproceedings{xu2022gmflow,
  title={Gmflow: Learning optical flow via global matching},
  author={Xu, Haofei and Zhang, Jing and Cai, Jianfei and Rezatofighi, Hamid and Tao, Dacheng},
  booktitle={Proceedings of the IEEE/CVF conference on computer vision and pattern recognition},
  pages={8121--8130},
  year={2022}
}

@article{guo2022review,
  title={A review of wavelet analysis and its applications: Challenges and opportunities},
  author={Guo, Tiantian and Zhang, Tongpo and Lim, Enggee and Lopez-Benitez, Miguel and Ma, Fei and Yu, Limin},
  journal={IEEe Access},
  volume={10},
  pages={58869--58903},
  year={2022},
  publisher={IEEE}
}

\end{document}